\documentclass[sigconf,screen]{acmart}
\AtBeginDocument{%
  }

\copyrightyear{2023}
\acmYear{2023}
\setcopyright{acmlicensed}\acmConference[MM '23]{Proceedings of the 31st ACM International Conference on Multimedia}{October 29-November 3, 2023}{Ottawa, ON, Canada}
\acmBooktitle{Proceedings of the 31st ACM International Conference on Multimedia (MM '23), October 29-November 3, 2023, Ottawa, ON, Canada}
\acmPrice{15.00}
\acmDOI{10.1145/3581783.3612368}
\acmISBN{979-8-4007-0108-5/23/10}

\usepackage[linesnumbered,ruled,vlined]{algorithm2e}
\usepackage{epsfig}
\usepackage{graphicx}
\usepackage{amsmath}

\usepackage{amssymb}
\usepackage{booktabs}
\usepackage{microtype}
\usepackage{pifont}
\usepackage{enumitem}
\usepackage{multirow}
\usepackage{makecell}
\usepackage{balance}
\newcommand{\cmark}{\ding{51}}
\newcommand{\xmark}{\ding{55}}

\begin{document}

\title{PoSynDA: Multi-Hypothesis Pose Synthesis Domain Adaptation for Robust 3D Human Pose Estimation}

\author{Hanbing Liu}
\authornote{Denotes equal contribution, authors are listed in random order}
\email{liuhb21@mails.tsinghua.edu.cn}
\affiliation{%
  \institution{Tsinghua University}
  \country{}
}

\author{Jun-Yan He}
\authornotemark[1]
\authornote{Zhi-Qi Cheng and Jun-Yan He are the corresponding authors}
\email{leyuan.hjy@alibaba-inc.com}
\affiliation{%
  \institution{DAMO Academy, Alibaba Group}
  \country{}
}

\author{Zhi-Qi Cheng}
\authornotemark[1]
\authornotemark[2]
\email{zhiqic@cs.cmu.edu}
\affiliation{%
  \institution{Carnegie Mellon University}
  \country{}
}

\author{Wangmeng Xiang}
\email{wangmeng.xwm@alibaba-inc.com}
\affiliation{%
  \institution{DAMO Academy, Alibaba Group}
  \country{}
}

\author{Qize Yang}
\email{qize.yqz@alibaba-inc.com}
\affiliation{%
  \institution{DAMO Academy, Alibaba Group}
  \country{}
}

\author{Wenhao Chai}
\email{wchai@uw.edu}
\affiliation{%
  \institution{University of Washington}
  \country{}
}

\author{Gaoang Wang}
\email{gaoangwang@intl.zju.edu.cn}
\affiliation{%
  \institution{Zhejiang University}
  \country{}
}

\author{Xu Bao}
\email{baoxu@email.szu.edu.cn}
\affiliation{%
  \institution{DAMO Academy, Alibaba Group}
  \country{}
}

\author{Bin Luo}
\email{luwu.lb@alibaba-inc.com}
\affiliation{%
  \institution{DAMO Academy, Alibaba Group}
  \country{}
}

\author{Yifeng Geng}
\email{cangyu.gyf@alibaba-inc.com}
\affiliation{%
  \institution{DAMO Academy, Alibaba Group}
  \country{}
}

\author{Xuansong Xie}
\email{xingtong.xxs@taobao.com}
\affiliation{%
  \institution{DAMO Academy, Alibaba Group}
  \country{}
}

\renewcommand{\shortauthors}{Hanbing Liu et al.}

\begin{abstract}
Existing 3D human pose estimators face challenges in adapting to new datasets due to the lack of 2D-3D pose pairs in training sets.
To overcome this issue, we propose \textit{Multi-Hypothesis \textbf{P}ose \textbf{Syn}thesis \textbf{D}omain \textbf{A}daptation} (\textbf{PoSynDA}) framework to bridge this data disparity gap in target domain.
Typically, PoSynDA uses a diffusion-inspired structure to simulate 3D pose distribution in the target domain.
By incorporating a multi-hypothesis network, PoSynDA generates diverse pose hypotheses and aligns them with the target domain.
To do this, it first utilizes target-specific source augmentation to obtain the target domain distribution data from the source domain by decoupling the scale and position parameters.
The process is then further refined through the teacher-student paradigm and low-rank adaptation.
With extensive comparison of benchmarks such as Human3.6M and MPI-INF-3DHP, PoSynDA demonstrates competitive performance, even comparable to the target-trained MixSTE model~\cite{zhang2022mixste}.
This work paves the way for the practical application of 3D human pose estimation in unseen domains.
The code is available at \hyperlink{blue}{https://github.com/hbing-l/PoSynDA}.
\end{abstract}



\maketitle

\vspace{-0.1in}
\section{Introduction}
The rise of meta-universes and various perception systems~\cite{zhao2018multi, huang2021generating, cheng2016video,cheng2017video,cheng2017video2shop,nguyen2017vireo,sun2018personalized,he2023damo, bao2023keyposs, cheng2022gsrformer,cheng2019learning, cheng2019improving, cheng2022rethinking}, has revved the need for advanced 3D Human Pose Estimation (3D HPE)~\cite{li20153d, zhou2016deep, sun2017compositional,zhou2022hypergraph, zhou2023overcoming, chen2023hdformer}.
Essential for various real-world applications, 3D HPE deals with the estimation of posture and temporal modeling within a 2D skeleton sequence.
Even with respectable advancements, prevalent strategies~\cite{he2021db,han2021transformer, dosovitskiy2020image, he2021transreid, ZhangCVPR22MixSTE, chen2023hdformer} still encounter difficulties due to the intricacies stemming from diverse scenes and the extensive range of 2D-3D human pose datasets~\cite{Ionescu_POSE_TPAMI14}. 
This, in turn, subsequently results in apparent challenges in achieving scenario adaptability.

\begin{figure*} [!ht]
	\begin{center}
		\includegraphics[width=0.95\linewidth]{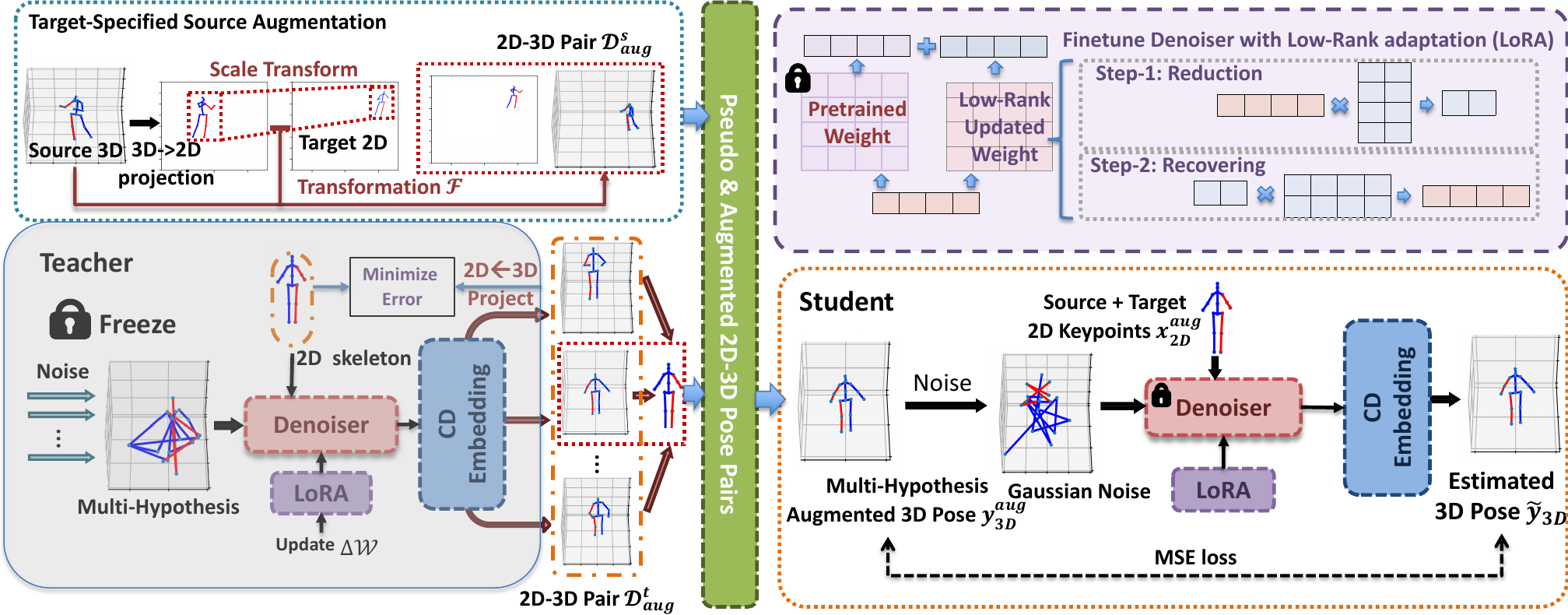}
	\end{center}
	\vspace{-0.15in}
\caption{\small The PoSynDA Framework. Augmented source 2D-3D pairs, $ \mathcal{D}_{aug}^s$, are derived through scale transformations on 2D skeletons, aligning them with the target 2D skeleton scale. The teacher network, with static parameters, generates multiple 3D pose hypotheses using noise samples and target 2D skeleton conditioning. These 3D poses are then projected to 2D, with the closest projection to the target 2D skeleton being chosen as its pseudo label, represented as $\mathcal{D}_{aug}^t$. In the student network, these augmented pairs are processed by a denoiser with LoRA and cross-dataset embedding to train the pose estimator $\mathcal{P}$ with parameter $\theta$.}
	\label{fig:pipeline}
	\vspace{-0.1in}
\end{figure*}

Therefore, gathering high-quality 2D-3D pose pairs in intricate target scenarios and addressing the dilemma of pose ambiguity and imprecise data transformations remain the paramount obstacles.
Previous efforts in domain adaptation for 3D HPE have been mainly focused on conventional augmentation schemes~\cite{2021PoseAug} and aligning source and target data within the same scale space~\cite{chai2023global}, often resulting in only marginal improvements. Motivated by recent progress in generative models~\cite{NEURIPS2020_4c5bcfec}, we introduce the \textit{Multi-Hypothesis \textbf{P}ose \textbf{Syn}thesis \textbf{D}omain \textbf{A}daptation} (PoSynDA) framework (Figure~\ref{fig:intro} and \ref{fig:pipeline}), focusing on the following five key aspects:

\begin{figure} [!ht]
	\begin{center}
		\includegraphics[width=0.8\linewidth]{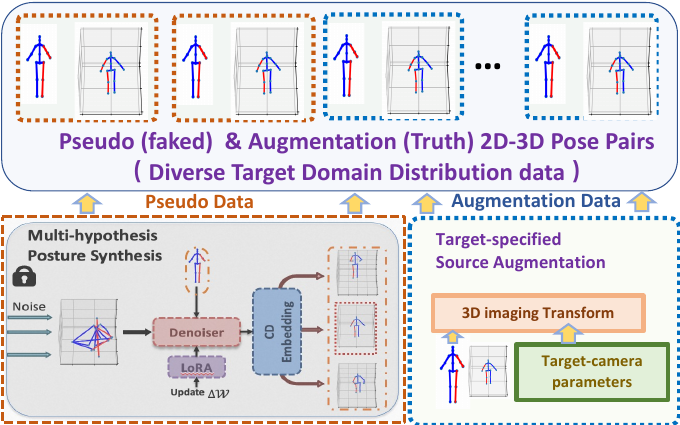}
	\end{center}
	\vspace{-0.15in}
\caption{\small Overview of Multi-Hypothesis Pose Synthesis. Multiple viable 3D poses for the target domain are generated using the target's 2D skeleton. The most accurate hypothesis is selected as the pseudo label. Target-specific source augmentation is applied, aligning generated 2D-3D pose pairs with the target domain's distribution.}
	\label{fig:intro}
	\vspace{-0.15in}
\end{figure}

\begin{enumerate}

\item \textbf{Data Synthesis}: \textit{Enhanced domain-adaptation} through a generative, target-specific source augmentation, using a multi-hypothesis way to emulate the target data distribution.

\item \textbf{Domain Alignment}: \textit{Replication of diverse target-domain data} by decoupling the scale factor from domain adaptation, aligning source and target data, and simplifying the underlying distribution.

\item \textbf{Multi-Hypothesis Testing}: \textit{Realistic data distribution} via a multi-hypothesis synthesis generative pipeline, synthesizing pseudo-data and re-projecting it into the target domain.

\item \textbf{Scenario Adaptability}: \textit{Effective optimization strategy} employing a teacher-student learning paradigm to conduct model training, using teacher network to generate multiple hypotheses and reduce memory usage, while guiding in the generalization ability of student network.

\item \textbf{Continuous Learning}: \textit{Efficient domain adaptation} with low-rank adaptation for large diffusion-based model fine-tuning, optimizing only a minimal set of parameters.
\end{enumerate}

\textit{In summary, our proposed PoSynDA presents a paradigm shift in domain adaptation for 3D Human Pose Estimation.} Our experiments show that it outperforms existing methods, achieving a 58.2mm MPJPE without using 3D labels from the target domain, comparable with the performance of the target-specific MixSTE model (58.2mm vs. 57.9mm)\cite{zhang2022mixste}. This work sets a new benchmark and opens new routes for exploration, extending to various perception-based interactions and meta-universe systems, with the potential to enhance their effectiveness and application range.

\section{Related Works}
\noindent \textbf{3D Human Pose Estimation.}~3D Human Pose Estimation (3D HPE) has attracted notable attention due to its applicability in various domains \cite{xiang2022language, xiang2022spatiotemporal, tu2023implicit,hauptmann2023robust, he2023damo,li2023longshortnet, lan2023procontext,wang2021deep}. The existing methods are broadly classified into 1) one-step, direct estimation of 3D poses, and 2) two-step, the elevation of 2D keypoints to 3D. Recent innovations include transformer-based approaches such as PoseFormer~\cite{zheng20213d}, MixSTE~\cite{zhang2022mixste}, and GCN~\cite{wang2020motion}. Additionally, several works addressed single-view 3D HPE by generating different hypotheses\cite{jahangiri2017generating, li2022mhformer, shan2023diffusion}. In contrast, our PoSynDA innovatively generates and selects the optimal postures as pseudo-labels during training, enriching the target domain data and adapting to unsupervised scenarios.

\noindent \textbf{Domain Adaptation in 3D HPE.}~The challenge of the domain gap in 3D HPE requires ingenious solutions. Generally, existing strategies include 1) cross-dataset adaptation such as BOA~\cite{Guan_Bilevel_CVPR21} and frame-by-frame optimization~\cite{Zhang_Infer_nips20}, and 2) data augmentation such as MoCap~\cite{MoCap_Rogez_NIPS16} and differentiable pose augmentation~\cite{2021PoseAug}. These works mark some progress in matching the data distribution of the target domain. However, our PoSynDA goes a step further by uniquely integrating generative techniques to mirror the target data distribution, overcoming the barriers in domain adaptation.

\noindent \textbf{Diffusion Models in Generative Tasks.}~Diffusion models such as DDPMs~\cite{NEURIPS2020_4c5bcfec} have opened new ways in generative tasks~\cite{poole2022dreamfusion, tevet2022human, xu2023odise}, deconstructing and reconstructing data in various applications, from image and 3D model generation to human motion generation and object detection.
Uniquely, our PoSynDA leverages the capabilities of diffusion models to directly synthesize high-fidelity 3D human poses, a distinction that sets it apart from existing approaches.
Our research addresses the inherent challenges of 3D human pose estimation.

\section{The PoSynDA Framework}
\subsection{Problem Definition}
In 3D Human Pose Estimation (3D HPE), the challenge of unsupervised domain adaptation is to predict accurate 3D human poses across varied conditions—such as differing lighting, camera angles, or motion dynamics—without the benefit of labeled data from the target domain. Given the high costs and time involved in collecting labeled 3D human pose data, domain adaptation becomes an essential tool. Typically, It can harness the labeled data from a recognized source domain, using it to fine-tune 3D human pose estimation in domains where such labeled data is missing.  In essence, domain adaptation seeks to generate a model that seamlessly bridges the data discrepancy between source and target domains, facilitating universal generalization across varying scenarios.

Formally, we define the source and target domains as $\mathcal{D}_s ={(\boldsymbol{x_{2D}^s},\boldsymbol{y_{3D}^s})}_{i=1}^{n_s}$ and $\mathcal{D}_t = {(\boldsymbol{x_{2D}^t})}_{i=1}^{n_t}$, respectively. The source domain $\mathcal{D}_s$ includes 2D human keypoints $\boldsymbol{x_{2D}^s}$ and corresponding 3D coordinates $\boldsymbol{y_{3D}^s}$, while the target domain $\mathcal{D}_t$ contains only 2D keypoints $\boldsymbol{x_{2D}^t}$. The objective is to design a 3D pose estimator $\mathcal{P}$ with parameter $\theta$ to convert 2D keypoints into 3D poses, leading to the following optimization problem:
\begin{equation}
\min_{\theta} \mathcal{L}_{\mathcal{P}}\left(\mathcal{P}_{\theta}, \mathcal{D})=\mathcal{L}_{\mathcal{P}}\left(\mathcal{P}_\theta(\boldsymbol{x_{2D}} \right), \boldsymbol{y_{3D}}\right),
\end{equation}
where $\mathcal{D}={ (\boldsymbol{x_{2D}}, \boldsymbol{y_{3D}}) }$ consists of paired 2D-3D poses, and the loss function $\mathcal{L}$ corresponds to the mean square errors (MSE) between predicted 3D poses and ground truths.

As shown in Figure~\ref{fig:intro}, we first initialize the 3D pose estimator for the target domain using parameters $ \theta_s $ trained on the source domain. Then, through data augmentation and pseudo-labeling, the model is adapted to the target domain. The augmented data pair $\mathcal{D}_{aug}$ is used to fine-tune $\mathcal{P}$, resulting in the optimization problem:
\begin{equation}
\min_{\theta} \mathcal{L}_{\mathcal{P}}\left(\mathcal{P}_{\theta}, \mathcal{D}_{aug}; \theta_s \right)=\mathcal{L}_{\mathcal{P}}\left(\mathcal{P}_{\theta} \left( \boldsymbol{x_{2D}^{aug}} \right), \boldsymbol{y_{3D}^{aug}} \right),
\label{eq:min}
\end{equation}
where $\mathcal{D}_{aug}=(\boldsymbol{x_{2D}^{aug}}, \boldsymbol{y_{3D}^{aug}})$ includes the augmented source data and pseudo-labeled target data. The ultimate goal is to train a function $\mathcal{P}$ that accurately predicts the 3D human pose $\boldsymbol{\widehat{y}_{3D}^{t}}$ in the target domain, using only the labeled source domain and the unlabeled target domain.

In general, as depicted in Figure~\ref{fig:pipeline}, our PoSynDA adopts a teacher-student paradigm within a diffusion model framework. Here, the denoiser acts as $\mathcal{P}$, converting 2D keypoints to 3D poses. Multiple 3D pose hypotheses are generated through repeated sampling Gaussian noise and lifted using denoiser $\mathcal{P}$, and the pose closest to the ground truth is chosen as the pseudo-label. This strategy helps provide supervision in the absence of real labels. PoSynDA further leverages 3D poses from the source domain and 2D keypoints from the target domain to create a unified dataset, effectively merging both domains. Specific estimation details and the entire model structure are discussed in the following sections.

\subsection{Target-specified Source Data Augmentation}
\label{sec:source_aug}
3D Human Pose Estimation (3D HPE) is often challenged by domain shift due to differences in camera intrinsic and extrinsic parameters across datasets. To overcome this issue and improve the performance of the target domain, we propose a target-specified source data augmentation strategy. By utilizing known camera parameters of the target dataset, we employ a 3D imaging algorithm to adapt the source data's 3D labels to the target domain. The focus of this scheme is to minimize the impact of scale and position variations, which can be detrimental during domain adaptation.

Consider a 3D pose from the source domain dataset $\mathcal{D}_s$, centered at the origin $[0, 0, 0]$, and a 2D pose from the target domain $\mathcal{D}_t$. Leveraging methods from previous work \cite{chai2023global}, we randomly sample a pose pair $(\boldsymbol{y_{3D}^s},\boldsymbol{x_{2D}^t})$ to approximate the target domain's scale and position distribution using Monte Carlo techniques \cite{hastings1970monte}. A transformation function $\mathcal{F}$ is then applied to convert the source 3D pose to the 2D target domain, as follows:
\begin{equation}
    \mathcal{D}_{aug}^{s} = \mathcal{F}(\boldsymbol{x_{2D}^t}, \boldsymbol{y_{3D}^s}),
\label{eq:aug_s}
\end{equation}
where $\mathcal{D}_{aug}^{s}$ signifies the transformed 3D pose aligned with the target domain's distribution. A comprehensive description of the transformation function $\mathcal{F}$ is found in \cite{chai2023global}. By performing this target-specified source data augmentation, we effectively bridge the gap between the source and target domains. This not only enhances the adaptability of the model but also leads to improved 3D human pose estimation performance across diverse conditions.

\subsection{Multi-hypothesis Domain Adaptation}
\label{sec:pseudo}
3D Human Pose Estimation (3D HPE) is complex due to depth ambiguity and self-occlusion, where multiple valid 3D solutions might correspond to a single 2D pose. To address this issue, PoSynDA synthesizes multiple plausible 3D poses, selecting the most likely one to represent the target. This probabilistic approach compensates for limited diversity in the target domain and lack of labeled data by utilizing multiple hypotheses to approximate the 3D pose distribution, adopting the best-matching pose as a pseudo-label. In the 2D-to-3D lifting model, the generation of multiple hypotheses is achieved by repeated sampling of noise from a standard Gaussian distribution. The number of hypotheses, denoted as $H$, balances accuracy and computational efficiency, increasing hypothesis space coverage and posing diversity as $H$ grows.

Specifically, PoSynDA utilizes a teacher-student learning paradigm, where both networks share identical weights but perform different roles. The teacher generates hypotheses and pseudo-labels without updating during training, while the student uses pseudo-labels to adjust their parameters.

\begin{itemize}
    \item \textbf{Teacher Network.}~The input 2D keypoints are transformed into a Gaussian distribution by gradually introducing noise. Subsequently, noise $\epsilon \sim \mathcal{N}(0, \boldsymbol{I})$ is sampled to restore the 3D pose via a denoiser. By sampling $H$ noises, we derive $H$ different poses, denoted by $\boldsymbol{\widetilde{y}}_{0:H}$ for each frame, which are then projected into the 2D camera plane. The pseudo-label is determined by calculating the error between these projections and the original 2D keypoints $\boldsymbol{x_{2D}^{t}}$, selecting the hypothesis with the minimum error as the pseudo-label:
    \begin{gather}
    h^{\prime}=\underset{h \in[0, H]}{\arg \min }\left\|\mathcal{G}\left(\boldsymbol{\widetilde{y}}_{h}\right)-\boldsymbol{x_{2D}^{t}}\right\|_2, \\
    \boldsymbol{\widetilde{y}_{3D}}=\boldsymbol{\widetilde{y}}_{h^{\prime}},
    \label{eq:hypo}
    \end{gather}
    where $\mathcal{G}$ is the projection function, and $\boldsymbol{\widetilde{y}_{3D}}$ is the pseudo-label. The teacher focuses solely on hypothesis generation and pseudo-label determination without participating in gradient computation.

    \item \textbf{Student Network.}~The student learns from both augmented source data $\mathcal{D}_{aug}^s$ (from Sec. \ref{sec:source_aug}) and target data $\mathcal{D}_{aug}^t$ obtained by the teacher. Unlike the teacher's multi-hypothesis approach, the student generates only a single 3D pose, signified by setting $H$ to 1. After each student updates using $\mathcal{D}_{aug}$, the corresponding parameters are synchronized back to the teacher, maintaining coherence throughout the learning process.
\end{itemize}

Note that the coordinated process is detailed in Algorithm \ref{algr:training pipeline}, encapsulating our multi-hypothesis domain adaptation. By leveraging both the teacher and student networks, it offers an efficient and robust solution to the challenges of 3D human pose estimation, particularly in cases with limited target domain diversity.

\begin{algorithm}[ht]
\DontPrintSemicolon
  \KwIn{Source domain $\mathcal{D}_s=(\boldsymbol{x_{2D}^s},\boldsymbol{y_{3D}^s})$, Target domain $\mathcal{D}_t=(\boldsymbol{x_{2D}^t})$, 3D Pose Estimator $\mathcal{P}$ with parameter $\theta$ initialized with $\theta_s$ trained on $\mathcal{D}_s$, Number of hypotheses $H$, Projection function $\mathcal{G}$, Loss function $\mathcal{L}$, learning rate $\eta$}
  \KwOut{Estimated pose $\boldsymbol{\widetilde{y}_{3D}^t}$ of target domain}
  \BlankLine
\textbf{Training:}\\
\While{$\theta$ has not converged}{
/* sampling batch data from datasets */ \\
Sample a batch from $\mathcal{D}^t=\{(\boldsymbol{x_{2D}^t})\}$  \\
Sample a batch from $\mathcal{D}^s=\{(\boldsymbol{x_{2D}^s},\boldsymbol{y_{3D}^s})\}$ \\
Augment the source data $ \mathcal{D}_{aug}^s=\mathcal{F}(\boldsymbol{x_{2D}^t}, \boldsymbol{y_{3D}^s})$ derived by Eq. \ref{eq:aug_s}\\
/* computing pseudo label for target data */ \\
\textbf{freeze} Teacher $\mathcal{P}$\\
\For{$h \leftarrow 0$ \KwTo $H$}{
sample noise $\epsilon \sim \mathcal{N}(0, \boldsymbol{I})$\\
$\boldsymbol{\widetilde{y}_h}=\mathcal{P}(\boldsymbol{x_{2D}^t},\epsilon)$ \\
}
$\boldsymbol{\widetilde{y}_{3D}}=\boldsymbol{\widetilde{y}_{h^{\prime}}}, h^{\prime}=\underset{h \in[0, H]}{\arg \min }\left\|\mathcal{G}\left(\boldsymbol{\widetilde{y}_{h}}\right)-\boldsymbol{x_{2D}^t}\right\|_2$ \\
$\mathcal{D}_{aug}^t=\{(\boldsymbol{x_{2D}^t},\boldsymbol{\widetilde{y}_{3D}})\}$\\
/* training the student estimator $\mathcal{P}$ */ \\
$\mathcal{D}_{aug}=concat(\mathcal{D}_{aug}^s,\mathcal{D}_{aug}^t)=\{(\boldsymbol{x_{2D}^{aug}}, \boldsymbol{y_{3D}^{aug}})\}$ \\
Sample noise $\epsilon \sim \mathcal{N}(0, \boldsymbol{I})$ and forward\\
Compute the loss and gradients by Eq.~\ref{eq:min}\\
Updating $\theta$ using Adam with 
$\theta \leftarrow \theta-\eta \nabla_\theta \mathcal{L}_{\mathcal{P}}(\mathcal{P}(\boldsymbol{x_{2D}^{aug}},\epsilon), \boldsymbol{y_{3D}^{aug}})$ \\
Update Teacher and Student $\mathcal{P}$ with $\theta$
}
\textbf{Inference:} \\
\textbf{freeze $\mathcal{P}$} \\
sample noise $\epsilon \sim \mathcal{N}(0, \boldsymbol{I})$ \\
$\boldsymbol{\widetilde{y}_{3D}^t}=\mathcal{P}(\boldsymbol{x_{2D}^t},\epsilon)$ \\
\caption{Domain adaptation training and inference algorithm}
\label{algr:training pipeline}
\end{algorithm}

\subsection{Model Structure}
\label{sec:model_struc}
Our proposed PoSynDA framework is designed around the diffusion model, consisting of a denoiser, LoRA (low-rank adaptation) module, and cross-dataset embedding components. These are depicted in Figure \ref{fig:pipeline}. Below we clarify the underlying principles of the diffusion model and describe each component in detail.

\noindent \textbf{Diffusion Model.}~The Denoising Diffusion Probabilistic Model (DDPM)~\cite{NEURIPS2020_4c5bcfec} serves as the core of the generative model, comprising two main processes. Firstly, a diffusion process progressively introduces Gaussian noise to the data. Secondly, a denoising process rebuilds the data from this noise utilizing a denoiser. Through iterative noise addition and denoising, the neural network (NN) learns to transform any Gaussian noise into the target data distribution.

Consider the target data $\boldsymbol{y}_0$. The forward process $q$ gradually incorporates Gaussian noise, $\epsilon$, with a variance of $\beta_{e} \in [0, 1]$ at each time step $e$. This leads from $\boldsymbol{y}_1$ to $\boldsymbol{y}_E$ as follows:
\begin{equation}
    q(\boldsymbol{y}_e|\boldsymbol{y}_{e-1}) = \mathcal{N}(\boldsymbol{y}_e; \sqrt{1 - \beta_{e}} \boldsymbol{y}_{e-1}, \beta_{e}\mathbf{I}),
\label{eq:ddpm1}
\end{equation}
where we use the Markov chain properties, and $\boldsymbol{y}_e$ in Equation~\ref{eq:ddpm1} can be sampled directly, only with $y_0$ as the condition, as:
\begin{equation}
    q(\boldsymbol{y}_e|\boldsymbol{y}_0) = \mathcal{N}(\boldsymbol{y}_e; \sqrt{\bar{\alpha}_e} \boldsymbol{y}_0, (1-\bar{\alpha}_e)\mathbf{I}),
\label{eq:ddpm2}
\end{equation}
where $\alpha_e = 1 - \beta_{e}$ and $\bar{\alpha}_e = \Pi^{e}_{s=1} \alpha_s$. For 3D human pose estimation, the noisy 3D pose $\boldsymbol{y_e}$ is fed to a denoiser $\mathcal{P}$, conditioned on 2D keypoints $\boldsymbol{x_{2D}}$ and time step $e$, to reconstruct the noise-free 3D pose $\widetilde{\boldsymbol{y}}_{0}$:
\begin{equation}
    \widetilde{\boldsymbol{y}}_{0}=\mathcal{P}(\boldsymbol{x}_{2D},\boldsymbol{y}_e,e).
\end{equation}
where $\widetilde{\boldsymbol{y}}_{0}$ represents the estimated pose $\widetilde{\boldsymbol{y}}_{3D}$. The denoising network is guided by a simple MSE loss:
\begin{equation}
\mathcal{L} = || \boldsymbol{y}_0 - \widetilde{\boldsymbol{y}}_0||_{2}.
\label{eq:loss}
\end{equation}

To create $H$ hypotheses $\boldsymbol{\widetilde{y}}_{0: H,0}$ in the target domain, we repeatedly sample from a Gaussian distribution, utilizing the concatenation of 3D Gaussian noise $\boldsymbol{\widetilde{y}}_{0: H,e}$ and $\boldsymbol{x}_{2D}^t$ as inputs for the denoiser $\mathcal{P}$. The optimal hypothesis is selected by choosing the estimated target pose with the minimal 2D projection error, as described in Equation~\ref{eq:hypo}. This elegant formulation underpins the robustness and efficiency of our approach to 3D human pose estimation, artfully handling both the diffusion and denoising processes.

\noindent \textbf{Denoiser Model.}~Our proposed PoSynD uniquely stands out by eliminating the need for designing a separate denoiser network structure, allowing seamless integration with existing 2D-to-3D human pose estimation networks. This ensures not only forward compatibility but also broad applicability across various domains. We strategically employ the cutting-edge MixSTE \cite{zhang2022mixste} as our denoiser, leveraging its proven effectiveness in 2D-to-3D pose estimation. Our decision reflects a carefully considered alignment with current state-of-the-art technologies, providing a robust foundation for our framework.

Additionally, in the experiment, PoSynD undertakes an in-depth evaluation to assess the feasibility of employing other 2D-to-3D lifting networks, such as VideoPose \cite{pavllo20193d}, as potential denoisers. This underscores our commitment to versatility and continuous innovation in optimizing the denoising strategy.

\noindent \textbf{Low-Rank Adaptation (LoRA).}~Full fine-tuning of larger denoisers that encompasses all model parameters is often cumbersome and inefficient. In contrast, PoSynDA leverages the low-rank adaptation (LoRA) technique \cite{hu2021lora} to facilitate a streamlined and cost-effective adaptation, targeting only essential components. While LoRA was originally conceived for large-scale language models within transformer blocks, we innovatively extend its application to 3D pose estimation.

In the 3D pose estimator $\mathcal{P}$, pre-trained on source data, we use query, key, value, and output projection matrices, represented as $W$. With $W_0$ denoting a pre-trained weight matrix and $\Delta W$ the gradient update during adaptation, the rank of the LoRA module is defined as $r$. The update to the weight matrix is thus constrained to a low-rank decomposition form, expressed as \(W_0+\Delta W = W_0 + BA\), where \(B \in \mathbb{R}^{d\times r}\), \(A \in \mathbb{R}^{r\times k}\), and the rank \(r \ll min(d,k)\). 

Typically, the LoRA approach allows us to keep $W_0$ fixed while training the low-rank components $A$ and $B$, thus formulating the forward pass of projection as:
\begin{equation}
\large
p = W_0 \boldsymbol{x} + \Delta W \boldsymbol{x} = W_0 \boldsymbol{x} + BA\boldsymbol{x},
\end{equation}
where $p$ denotes the resultant hidden state, and $\boldsymbol{x}$ represents the input queries or tokens. With proper initialization for matrices $A$ and $B$, $\Delta W = BA$ starts as zero, and is gradually adjusted during adaptation. Our experiments primarily use a rank $r$ of 4, with variations tailored to specific scenarios. This strategic use of LoRA emphasizes efficiency without sacrificing accuracy, demonstrating our commitment to state-of-the-art adaptation techniques for 3D pose estimation.

\begin{table}[!t]
\centering
\setlength{\tabcolsep}{1mm}
\begin{tabular}{l|c|cc}
    \specialrule{1pt}{1pt}{2pt}
    Method & S & MPJPE~($\downarrow$) & P-MPJPE~($\downarrow$)\\
    \hline
    Pavllo~\textit{et al.}~\cite{pavllo20193d} & Full & 37.2 & 27.2 \\
    Cai~\textit{et al.}~\cite{cai2019exploiting} & Full & 50.6 &  40.2\\
    Martinez~\textit{et al.}~\cite{martinez2017simple} & Full & 45.5 & 37.1 \\
    Zhao~\textit{et al.}~\cite{zhao2019semantic} & Full & 43.8 & -\\
    Lui~\textit{et al.}~\cite{liu2020attention} & Full & 34.7 & - \\
    Wang~\textit{et al.}~\cite{wang2020motion} & Full & 25.6 & - \\
    \hline
    Li~\textit{et al.}~\cite{li2020cascaded} & S1 & 50.5 & - \\
    Pavllo~\textit{et al.}~\cite{pavllo20193d} & S1 & 51.7 & - \\
    Gong~\textit{et al.}~\cite{2021PoseAug} & S1 & 56.7 & -\\
    Gholami~\textit{et al.}~\cite{gholami2022adaptpose} & S1 & 54.2 & 35.6\\
    Chai~\textit{et al.}~\cite{chai2023global} & S1 & 49.9 & 34.2\\
    \hline
    \textbf{\textit{Ours}}  & S1 & \textbf{48.1} & \textbf{33.2}\\
    \specialrule{1pt}{1pt}{2pt}
\end{tabular}
\vspace{1.5mm}
\caption{\small \textbf{Quantitative Results on H3.6M.} S represents the source domain. MPJPE and P-MPJPE are used as evaluation metrics. Source: S1. Target: S5, S6, S7, S8.}
\label{tab:H3.6M}
\vspace{-0.2in}
\end{table}

\noindent \textbf{Cross-Dataset Embedding.}~Our proposed PoSynDA introduces a targeted strategy to mitigate biases and inconsistencies in the 3D pose estimator trained on source data, aligning with research directions found in works like \cite{zaken2021bitfit,cai2020tinytl}. The final goal is to eliminate the disparities in scale and position that might manifest between the source and target domains, thereby creating a more versatile and robust estimator.

Specifically, PoSynDA fulfills this objective by incorporating an additional embedding layer at the output stage of the estimator $\mathcal{P}$. Specifically, during adaptation or inference, for any given condition $\boldsymbol{x_{2D}}$, whether derived from augmented source or target data, we integrate a bias into the predicted 3D pose to produce the finalized pose estimation, symbolized as $\mathcal{P}_{\theta}(\boldsymbol{x_{2D}},\epsilon)$. The procedure is mathematically expressed as:
\begin{equation}
\mathcal{P}_{\theta}(\boldsymbol{x_{2D}},\epsilon) = \boldsymbol{\bar{y}_{3D}} + B_{\text{bias}},
\end{equation}
where $\boldsymbol{\bar{y}_{3D}}$ denotes the intermediate output of $\mathcal{P}$, and $B_{\text{bias}}\in \mathbb{R} ^{3\times J}$ is formulated through integration of designated embedding layer.

By employing this cross-dataset embedding, PoSynDA bridges the gap between the source and target domains, ensuring that the predicted poses conform more closely to the expectations of the target context. This innovation enhances the estimator's precision and adaptability, fostering greater accuracy and reliability in 3D pose estimation across various datasets.

\begin{table}[!t]
\centering
\setlength{\tabcolsep}{1mm}
\begin{tabular}{l|c|ccc}
    \specialrule{1pt}{1pt}{2pt}
    Method & CD & MPJPE~($\downarrow$) & PCK~($\uparrow$) & AUC~($\uparrow$) \\
    \hline
    Mehta~\textit{et al.}~\cite{Mehta_RCFSXT_3DV17} & & 117.6 & 76.5 & 40.8 \\
    VNect~\cite{mehta2017-vnect}  & & 124.7 & 76.6 & 40.4 \\
    OriNet~\cite{luo2018orinet} & & 89.4 & 81.8 & 45.2  \\
    Multi Person~\cite{mehta2018single} & & 122.2 & 75.2 & 37.8 \\
    Martinez~\textit{et al.}~\cite{martinez2017simple} & & 84.3 & 85.0 & 52.0\\
    Zhang~\textit{et al.}~\cite{zhang2022mixste} & & 57.9 & 94.2 & 63.8 \\
    \hline	
    Guan~\textit{et al.}~\cite{guan2021bilevel} & \checkmark & 117.6 & 90.3 & - \\
    Kanazawa~\textit{et al.}~\cite{kanazawa2018end} & \checkmark & 113.2 & 77.1 & 40.7\\
    Wandt~\textit{et al.}~\cite{wandt2019repnet} & \checkmark & 92.5 & 81.8 & 54.8 \\
    Ci~\textit{et al.}~\cite{ci2019optimizing} & \checkmark & - & 74.0 & 36.7\\
    Zeng~\textit{et al.}~\cite{zeng2020srnet} & \checkmark & - & 77.6 & 43.8\\
    Li~\textit{et al.}~\cite{li2020cascaded} & \checkmark & 99.7 & 81.2 & 46.1 \\
    Gong~\textit{et al.}~\cite{2021PoseAug}  & \checkmark & 92.6 & 82.9 & 46.5\\
    Gholami~\textit{et al.}~\cite{gholami2022adaptpose} & \checkmark & 77.2 & 88.4 & 54.2\\
    Chai~\textit{et al.}~\cite{chai2023global}  & \checkmark & 61.3& 92.1 & \textbf{62.5} \\
    \hline
    \textbf{\textit{Ours (VideoPose)}}  & \checkmark & 60.2 & 93.1 & 58.4 \\
    \textbf{\textit{Ours (MixSTE)}}  & \checkmark & \textbf{58.2}& \textbf{93.5} & 59.6 \\
    \specialrule{1pt}{1pt}{2pt}
\end{tabular}
\vspace{1.5mm}
\caption{\small \textbf{Quantitative Results on 3DHP}. CD refers to cross-domain evaluation, while no CD denotes fully supervised learning on the target domain. PCK, AUC, and MPJPE are used as evaluation metrics. Source: H3.6M. Target: 3DHP.}
\label{tab:3dhp}
\vspace{-0.2in}
\end{table}

\begin{table*}
    \centering
    \setlength{\tabcolsep}{1.5mm}{
    \begin{tabular}{l|c c c c c | c c c}
    \specialrule{1pt}{1pt}{2pt}
        Method & Denoiser & Source Data Augmentation & LoRA & CD Embedding & Multi-hypothesis & MPJPE[↓] & Params (K) & FLOPs (G) \\
    \hline
    Baseline  &  \cmark & \xmark &  \xmark &  \xmark &  \xmark & 111.7 & - & 277.26 \\
              & \cmark &  \cmark  & \xmark &  \xmark &  \xmark & 62.5 & - & 277.26 \\
              & \cmark &  \cmark  & \cmark &  \xmark &  \xmark & 61.9 & 196.60 & 278.88 \\
              & \cmark &  \cmark  & \cmark &  \cmark &  \xmark & 60.7 & 196.65 & 278.88 \\
    \hline
    \textbf{Ours}   & \cmark &  \cmark  & \cmark & \cmark & \cmark & \textbf{58.2} & \textbf{196.65} & \textbf{836.65} \\
\specialrule{1pt}{1pt}{2pt}
    \end{tabular}}
    \vspace{1.5mm}
\caption{\small Ablation study for each component in our method. The evaluation results are reported by MPJPE (mm), Parameters (K), and FLOPs (G). Source: H3.6M. Target: 3DHP.}
\label{table:ablation_component}
\vspace{-4mm}
\end{table*}

\section{Experiments}

\subsection{Datasets and Metrics}
Experiments are conducted on three 3D pose estimation datasets, each offering unique characteristics and challenges: Human3.6M (H3.6M) \cite{Ionescu_POSE_TPAMI14}, MPI-INF-3DHP (3DHP) \cite{mehta2017-vnect}, and 3DPW \cite{von2018recovering}.

\noindent \textbf{Human3.6m (H3.6M).}~H3.6M is recognized as one of the most extensive datasets for human pose estimation, comprising 3.6 million frames. It captures 11 subjects engaged in 15 diverse activities, such as walking, sitting, and jumping, under varying camera angles and lighting conditions. It includes detailed ground-truth annotations of skeletal joints, RGB videos, and camera calibration parameters. According to established works \cite{gholami2022adaptpose, pavllo20193d}, evaluations are conducted under two different setups. These setups encapsulate varying scenarios and complexities, reflecting the model's adaptability across different domains. The Mean Per Joint Position Error (MPJPE) and Procrustes-aligned Mean Per Joint Position Error (P-MPJPE) metrics are utilized for quantitative evaluation.

\noindent \textbf{MPI-INF-3DHP (3DHP).}~3DHP presents a vast and versatile dataset, including both indoor and outdoor scenes. Comprising 2,929 frames in the test set, the dataset adds complexity with its diverse backgrounds and environmental conditions. Evaluation on this dataset is multifaceted, employing metrics like MPJPE, Percentage of Correct Keypoints (PCK) with a 150mm threshold, and the Area Under the Curve (AUC) calculated across various PCK thresholds. These metrics provide a comprehensive understanding of the performance, especially in complex real-world scenarios.

\noindent \textbf{3D People in the Wild (3DPW).}~3DPW dataset provides a one-of-a-kind benchmark for 3D human pose estimation in uncontrolled outdoor settings. Captured with dynamic camera motion and natural lighting, 3DPW introduces variability in viewpoints, human positions, and imaging conditions. With more diverse camera angles compared to 3DHP and H3.6M, 3DPW evaluates methods at 25fps, adding temporal complexity. Models trained on indoor datasets like H3.6M are tested on the 3DPW test set using MPJPE and Procrustes-aligned MPJPE (P-MPJPE) metrics. By requiring generalization to complex in-the-wild scenarios, 3DPW pushes progress in robust 3D human pose estimation under poor imaging conditions and uncontrolled dynamics.

\vspace{-1mm}
\subsection{Implementation Details}
We implemented PoSynDA in PyTorch and performed training and inference leveraging the computational power of an NVIDIA TITAN V100 GPU. The Adam optimizer \cite{kingma2014adam} was employed with a tuned learning rate of 6e-5 to enable stable convergence. A batch size of 4 and 1,000 update steps were used during training. These hyperparameters were carefully selected through ablation studies to optimize performance across datasets. The input sequence lengths were set to 243 for Human3.6M, and 27 for 3DHP and 3DPW, following the settings in \cite{ZhangCVPR22MixSTE}. These values adequately capture the complexity of human motions in each dataset. To maximize training data diversity and strengthen generalization, we used a non-overlapping stride sampling approach with intervals equal to the input length. Through computational optimization and tuned hyperparameter selection, our implementation aims to effectively showcase the capabilities of PoSynDA.

\begin{table}[!t]
\small
\centering
\setlength{\tabcolsep}{1mm}

\begin{tabular}{l|c|cc}
    \specialrule{1pt}{1pt}{2pt}
    Method & CD  & P-MPJPE~($\downarrow$) & MPJPE~($\downarrow$)\\
    \hline
    Pavllo~\textit{et al.}~\cite{pavllo20193d} &  & 68.0 & 105.0\\
    Kocabas~\textit{et al.}~\cite{kocabas2020vibe} &  & 51.9 & 82.9\\
    Joo~\textit{et al.}~\cite{joo2021exemplar} &  & 55.7 & -\\
    Lin~\textit{et al.}~\cite{lin2021mesh} &  & 45.6 & 74.7\\
    \hline
    Kocabas~\textit{et al.}~\cite{kocabas2020vibe} & \checkmark & 56.5 & 93.5\\
    Gong~\textit{et al.}~\cite{2021PoseAug} & \checkmark & 58.5 & 94.1\\
    Guan~\textit{et al.}~\cite{guan2021bilevel} & \checkmark & 49.5 & 77.2\\
    Gholami~\textit{et al.}~\cite{gholami2022adaptpose} &\checkmark &46.5 & 81.2\\
    Chai~\textit{et al.}~\cite{chai2023global} & \checkmark  & 55.3 & 87.7\\
    \hline
    \textbf{\textit{Ours}} & \checkmark & \textbf{45.4} & \textbf{75.5}\\
    \specialrule{1pt}{1pt}{2pt}
\end{tabular}
\vspace{1.5mm}
\caption{\small \textbf{Quantitative Results on 3DPW}. CD refers to cross-domain evaluation, while no CD denotes fully supervised learning on the target domain. P-MPJPE and MPJPE are used as evaluation metrics. Source: H3.6M. Target: 3DPW.}
\label{tab:3dpw}
\vspace{-0.5in}
\end{table}

\vspace{-1mm}
\subsection{Quantitative Evaluation}
\noindent \textbf{Results on H3.6M.}~Table \ref{tab:H3.6M} presents a comparative analysis of our PoSynDA with previous unsupervised methodologies, focusing on those utilizing labeled source 2D-3D pairs and target 2D keypoints, while abstaining from using ground truth 3D poses. The table's upper section displays results utilizing the entirety of the H3.6M dataset as the source, whereas the lower section restricts the source data to subject S1. The target data spans subjects S5 to S8. By employing ground truth 2D keypoints, we ensure consistent evaluation conditions aligned with extant literature. However, it's pertinent to mention that our evaluations vary from some prior works due to the distinction between input videos and individual frames. Impressively, PoSynDA sets new benchmarks, demonstrating enhancements of 1.8mm and 1.0mm in MPJPE and P-MPJPE, respectively, over the prior methods.

\noindent \textbf{Results on 3DHP.}~Table \ref{tab:3dhp} summarizes the performance of PoSynDA on the 3DHP benchmark across multiple metrics including MPJPE, PCK, and AUC. This comprehensive analysis demonstrates our approach's efficacy for cross-dataset generalization. PoSynDA establishes state-of-the-art results on MPJPE and PCK, while achieving competitive AUC scores just behind PoseDA. Compared to prior best methods, PoSynDA reduces MPJPE by an impressive 3.1mm. Remarkably, when coupled with denoising techniques like VideoPose or MixSTE, our PoSynDA maintains top-ranked performance.

\noindent \textbf{Results on 3DPW.}~As shown in Table \ref{tab:3dpw}, our proposed PoSynDA achieves remarkable improvements over prior cross-dataset approaches on the 3DPW benchmark. Specifically, PoSynDA surpasses state-of-the-art methods by a sizeable margin of 1.7mm in MPJPE, thereby establishing new performance records. This demonstrates our method's ability to generalize effectively to complex in-the-wild 3D pose estimation, even when trained only on indoor datasets like Human3.6M. The significant gains unlocked by PoSynDA underscore its efficacy for unsupervised cross-domain 3D human pose estimation on challenging real-world benchmarks.

\begin{figure*}[!t]
    \centering
    \centerline{\includegraphics[width=\linewidth]{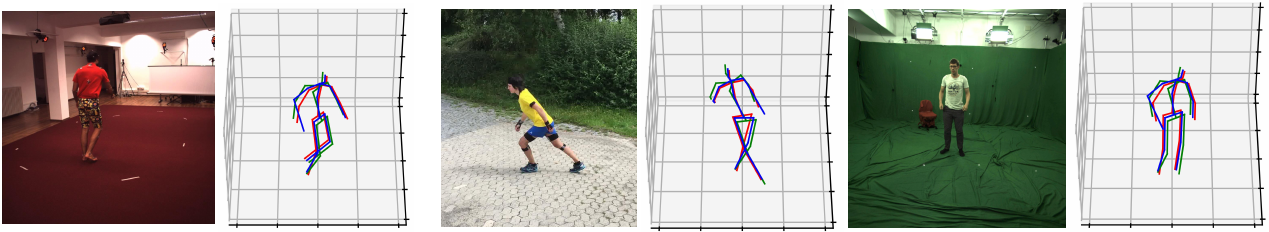}}
    \caption{\small \textbf{Multiple hypotheses generated by our method.} Each color represents a single hypothesis, and the red pose is selected as the pseudo label. These hypotheses showcase the diversity and rationality of the generated postures.}
    \label{fig:multi-h}
\end{figure*}

\begin{figure}[!t]
    \centering
    \centerline{\includegraphics[width=0.9\linewidth]{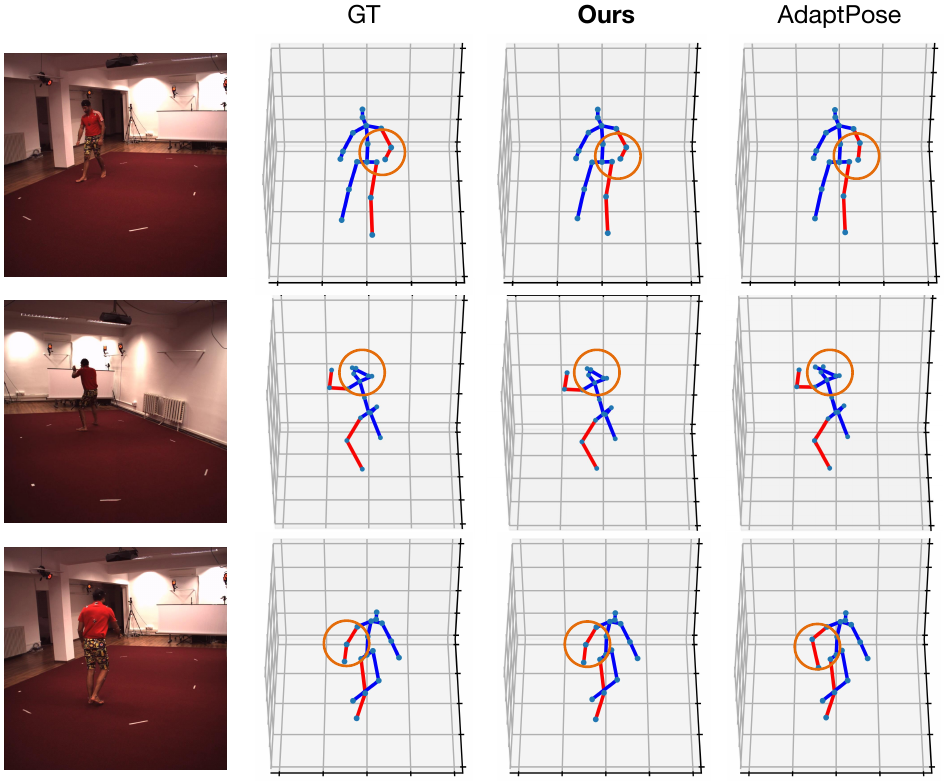}}
   \caption{\small \textbf{Comparison of 3D estimated human pose generated by different methods.} The figure displays the 3D reconstruction visualization results using our proposed method, the state-of-the-art method AdaptPose, ground truth, and the corresponding video frame from the Human3.6M dataset. The source domain is S1, and the target domains are S5, S6, S7, and S8. Our method exhibits higher accuracy and robustness in handling various actions and occlusion scenarios.}
    \vspace{-4mm}
    \label{fig:visualization}
\end{figure}

\subsection{Qualitative Evaluation}
\noindent \textbf{3D Reconstruction Visualization.}~We offer a detailed qualitative analysis of the H3.6M dataset, as demonstrated in Figure \ref{fig:visualization}. Our method's efficacy is thoroughly examined against renowned benchmarks such as AdaptPose \cite{gholami2022adaptpose}. Impressively, PoSynDA exhibits an outstanding ability to reconstruct both elementary and more intricate actions, even those obscured or occluded. This performance attests to our approach's robustness and stability, manifesting a nuanced understanding of human pose dynamics.

\noindent \textbf{Multi-Hypothesis Generation.}~A unique capability of our proposed PoSynDA is generating multiple plausible 3D pose hypotheses, as investigated in Figure \ref{fig:multi-h} across H3.6M, 3DHP, and 3DPW datasets. The visualized hypotheses are not only anatomically feasible and diverse, but also showcase precise alignment between the selected pose (in red) and the true underlying pose. This goes beyond visual concurrence - it reflects an understanding of the core principles governing human movement and behavior. Thus, it demonstrates both the technical excellence of PoSynDA's multi-hypothesis modeling and also the human-centric philosophy underpinning our approach. While quantitatively chosen based on the input image, the selected hypothesis visually resonates with the ground truth pose based on implicit knowledge of natural human poses. 

\begin{table}[!t]
\centering
\setlength{\tabcolsep}{1mm}
\begin{tabular}{c|c}
    \specialrule{1pt}{1pt}{2pt}
    Method & MPJPE~($\downarrow$)  \\
    \hline
    Multi-hypothesis Method: D3DP \cite{shan2023diffusion} & 96.5 \\
    Generative Method: GAN \cite{goodfellow2014generative} & 87.5 \\
    \hline
    \textbf{Ours} & \textbf{58.2} \\
    \specialrule{1pt}{1pt}{2pt}
\end{tabular}
\vspace{1.5mm}
\caption{\small \textbf{Ablation study:} The table compares our method's MPJPE performance with multi-hypothesis (D3DP) and generative (GAN) models. Source: H3.6M. Target: 3DHP.}
\label{tab:ablation_generative}
\vspace{-6mm}
\end{table}

\subsection{Ablation Studies}
As shown in Table \ref{table:ablation_component}, we conducted an ablation study on H3.6M as the source and 3DHP as the target domain. We started with a baseline model that was trained as a denoiser solely on the source domain, and then systematically integrated various components to analyze their effects.

\noindent \textbf{Baseline Configuration.}~The poor performance of the baseline model highlights the inherent complexity of unsupervised cross-domain 3D human pose estimation. By exposing the significant domain gap, the baseline results underscore the value of each component subsequently added to our PoSynDA approach.

\noindent \textbf{Source Data Augmentation.}~With the implementation of the source data augmentation module, the model experienced a remarkable 44.7\% improvement in MPJPE, emphasizing the pivotal role this component plays in enhancing prediction accuracy.

\noindent \textbf{LoRA Integration.}~The inclusion of LoRA further augmented the model's capabilities, marking its debut success in 3D human pose estimation. This achievement represents an innovative leap, unveiling LoRA's latent potential, and underscoring the significance of this novel integration within the framework of 3D pose analysis.

\noindent \textbf{Cross-Dataset Embedding.}~The incorporation of Cross-Dataset embedding, with minimal addition of only 0.05K parameters, succeeded in mitigating the bias between domains, strengthening the overall model performance. This efficient improvement illustrates the potential of nuanced optimization in enhancing the system's efficacy.

\noindent \textbf{Multi-Hypothesis Incorporation.}~Although the Multi-hypothesis module increased FLOPs substantially, it fine-tuned MPJPE by 2.5mm without affecting the inference speed, as it is employed solely within the teacher network during training. This highlights its distinct contribution to the model, emphasizing the strategic importance of this component within the architectural design.

\noindent \textbf{Comparison with Alternative Approaches.}~Our analysis also included scrutiny of other pioneering techniques, such as the Multi-hypothesis Method (D3DP) and GAN-based Generation Method. As shown in Table \ref{tab:ablation_generative}, these comparative evaluations, conducted using H36M as the source and 3DHP as the target, reveal that PoSynDA's performance surpasses both the latest state-of-the-art multi-hypothesis method, D3DP \cite{shan2023diffusion}, and the traditional GAN approach \cite{goodfellow2014generative}. This assessment further buttresses PoSynDA's standing as an innovative solution in the field of 3D pose estimation.

\begin{table}[!t]
\centering
\setlength{\tabcolsep}{1mm}
\begin{tabular}{c|ccc}
    \specialrule{1pt}{1pt}{2pt}
    \# of hypotheses & MPJPE~($\downarrow$) & PCK~($\uparrow$) & AUC~($\uparrow$) \\
    \hline
    1 & 60.7  & 92.5  & 58.2  \\
    2 & 59.9  & 93.4  & 59.1  \\
    3 & 58.2  & 93.5  & 59.6  \\
    4 & 58.1  & 93.7  & 59.3  \\
    5 & 58.3  & 93.7  & 59.4  \\
    6 & 58.5  & 93.4  & 58.9  \\
    \specialrule{1pt}{1pt}{2pt}
\end{tabular}
\vspace{1.5mm}
\caption{\small \textbf{Ablation study:} The table shows the impact of the number of hypotheses on the evaluation metrics MPJPE (mm), PCK (Percentage of Correct Keypoints), and AUC (Area Under the Curve). Source: H3.6M. Target: 3DHP.}
\label{tab:ablation_h}
\vspace{-2mm}
\end{table}

\begin{table}[!t]
\centering
\setlength{\tabcolsep}{1mm}
\begin{tabular}{c|ccc}
    \specialrule{1pt}{1pt}{2pt}
    Rank & MPJPE~($\downarrow$) & Params (K) & FLOPs (G)  \\
    \hline
    1 & 62.1 & 49.19 & 835.43 \\
    2 & 60.4 & 98.34 & 835.83 \\
    3 & 59.8 & 147.49 & 836.24 \\
    4 & 58.2 & 196.65 & 836.65 \\
    5 & 58.3 & 245.80 & 837.05 \\
    6 & 58.1 & 249.95 & 837.46 \\
    \specialrule{1pt}{1pt}{2pt}
\end{tabular}
\vspace{1.5mm}
\caption{\small \textbf{Ablation study:} The table presents the impact of different ranks on the diffusion model's performance, measured by MPJPE (mm), Parameters (K), and FLOPs (G). Source: H3.6M. Target: 3DHP.}
\label{tab:complexity}
\vspace{-5mm}
\end{table}

\noindent \textbf{Number of Hypotheses}~The influence of the number of hypotheses on the model's performance was systematically examined, with H3.6M as the source domain and 3DHP as the target domain. As presented in Table \ref{tab:ablation_h}, the experiments disclose a trend where increasing the number of hypotheses leads to an enhancement in the model's accuracy. Nevertheless, beyond the three hypotheses, further increments fail to yield significant gains; instead, they result in fluctuations within a specific range. Considering that the FLOPs of the model increase linearly with the number of hypotheses, our experimental configuration judiciously selected three hypotheses. This choice harmonizes the dual objectives of achieving high accuracy and maintaining computational efficiency.

\noindent \textbf{Structure of Diffusion Model}.~The nuanced interplay between the number of timesteps and the embedding dimension of the denoiser is explored in Table \ref{tab:ablation_timestap}. It reveals that augmenting both timesteps and embedding dimensions typically conduces to better performance. The zenith is reached at 1000 timesteps with an embedding dimension of 512, garnering the lowest MPJPE score of 58.2mm. Remarkably, further amplifying the embedding dimension to 1024 does not reap additional benefits, thus delineating an optimal configuration at 1000 timesteps and an embedding dimension of 512.

\noindent \textbf{Computational Complexity of LoRA}.~The LoRA component's computational complexity is intricately tied to its rank setting, introducing a multi-faceted trade-off between rank, computational overhead, and prediction accuracy within the realm of low-rank approximation. As Table \ref{tab:complexity} elucidates, our investigations identified a rank of 4 as a judicious selection within the LoRA architecture. This choice orchestrates an adept equilibrium between computational economy and accuracy, without sacrificing the integrity of the underlying mathematical construct.

\begin{table}[!t]
\centering
\setlength{\tabcolsep}{1mm}
\begin{tabular}{c|c|c}
    \specialrule{1pt}{1pt}{2pt}
    \# of timesteps & Dimension & MPJPE~($\downarrow$) \\
    \hline
    100 &  512  & 59.8  \\
    500 &  256  & 59.3  \\
    500 & 512   & 59.0  \\
    1000 & 256  &  58.9 \\
    1000 & 512  & 58.2  \\
    1000 & 1024 & 58.3  \\
    \specialrule{1pt}{1pt}{2pt}
\end{tabular}
\vspace{1.5mm}
\caption{\small \textbf{Ablation study:} The table shows the impact of different combinations of timesteps and denoiser embedding dimensions on MPJPE (Mean Per Joint Position Error). Source: H3.6M. Target: 3DHP.}
\label{tab:ablation_timestap}
\vspace{-4mm}
\end{table}

\section{Conclusion}
This paper presents PoSynDA, a framework for 3D human pose estimation using domain adaptation through multi-hypothesis pose synthesis. PoSynDA aims to generate a wide array of 3D poses in the target domain, addressing the challenge of limited diversity. PoSynDA operates using a diffusion-based structure, viewing pose estimation as a multi-step denoising diffusion process. Additionally, we propose a target-specified source augmentation scheme to create 3D pose pairs, adjusting for scale. Through evaluation of 3 benchmark datasets and comparison with state-of-the-art models, PoSynDA not only surpasses leading models but also competes with the target-domain trained model MixSTE \cite{zhang2022mixste}. Our future work will concentrate on real-world applications such as vision perception-based interaction and video generation to maximize the benefits of our proposed adaptation technique.

\section*{Acknowledgments}
The contributions of Zhi-Qi Cheng in this project were supported by the Army Research Laboratory (W911NF-17-5-0003), the Air Force Research Laboratory (FA8750-19-2-0200), the U.S. Department of Commerce, National Institute of Standards and Technology (60NANB17D156), the Intelligence Advanced Research Projects Activity (D17PC00340), and the US Department of Transportation (69A3551747111). Additionally, the Intel and IBM Fellowships also supported Zhi-Qi Cheng's research work. The views and conclusions contained herein represent those of the authors and not necessarily the official policies or endorsements of the supporting agencies or the U.S. Government.

\balance
\bibliographystyle{ACM-Reference-Format}
\bibliography{ref}

\end{document}